\documentclass[conference]{IEEEtran}
\IEEEoverridecommandlockouts
\usepackage{hyperref}
\usepackage{cite}
\usepackage{amsmath,amssymb,amsfonts}
\usepackage{algorithmic}
\usepackage{graphicx}
\usepackage{textcomp}
\usepackage{booktabs}
\usepackage{multirow}
\usepackage{xcolor}
\usepackage{caption} 
\usepackage{subcaption}
\usepackage{float}
\usepackage{array} 
\usepackage[ruled,vlined]{algorithm2e}

\usepackage{booktabs} 
\usepackage{adjustbox} 

\begin{document}

\title{BEFL: Balancing Energy Consumption in Federated Learning for Mobile Edge IoT}

\author{
    \IEEEauthorblockN{Zehao Ju\textsuperscript{1}, Fuke Shen\textsuperscript{2}, Tongquan Wei\textsuperscript{3,*}}
    \IEEEauthorblockA{\textit{Department of Computer Science and Technology, East China Normal University} \\
    Shanghai, China }
}
\maketitle

\begin{abstract}
Federated Learning (FL) is a privacy-preserving distributed learning paradigm designed to build a highly accurate global model. In Mobile Edge IoT (MEIoT), the training and communication processes can significantly deplete the limited battery resources of devices. Existing research primarily focuses on reducing overall energy consumption, but this may inadvertently create energy consumption imbalances, leading to the premature dropout of energy-sensitive devices.To address these challenges, we propose BEFL, a joint optimization framework aimed at balancing three objectives: enhancing global model accuracy, minimizing total energy consumption, and reducing energy usage disparities among devices. First, taking into account the communication constraints of MEIoT and the heterogeneity of devices, we employed the Sequential Least Squares Programming (SLSQP) algorithm for the rational allocation of communication resources. Based on this, we introduce a heuristic client selection algorithm that combines cluster partitioning with utility-driven approaches to alleviate both the total energy consumption of all devices and the discrepancies in energy usage.Furthermore, we utilize the proposed heuristic client selection algorithm as a template for offline imitation learning during pre-training, while adopting a ranking-based reinforcement learning approach online to further boost training efficiency. Our experiments reveal that BEFL improves global model accuracy by 1.6\%, reduces energy consumption variance by 72.7\%, and lowers total energy consumption by 28.2\% compared to existing methods. The relevant code can be found at \href{URL}{https://github.com/juzehao/BEFL}.

\end{abstract}

\section{Introduction}
\indent Federated Learning (FL) is a distributed and iterative machine learning approach that builds a global model while preserving user privacy by not requiring access to raw data \cite{ref3, ref4}. However, deploying FL in Mobile Edge IoT (MEIoT) environments introduces significant energy consumption challenges. Many IoT devices are battery-powered \cite{ref5}, which restricts their energy capacity. Consequently, energy optimization in FL has become a focal point for researchers.

For example, Kang et al. \cite{ref7} proposed the "Neurosurgeon" scheduler, which offloads parts of deep neural network (DNN) computations from mobile devices to data centers, reducing both latency and energy consumption in industrial IoT networks. Zhang et al. \cite{ref8} applied techniques like filter pruning and multitask learning to compress DNN models, lowering energy usage. However, these approaches often fail to consider the device heterogeneity present in MEIoT environments, which significantly affects energy consumption optimization.

\begin{figure}[t]
\centering
\includegraphics[width=\columnwidth]{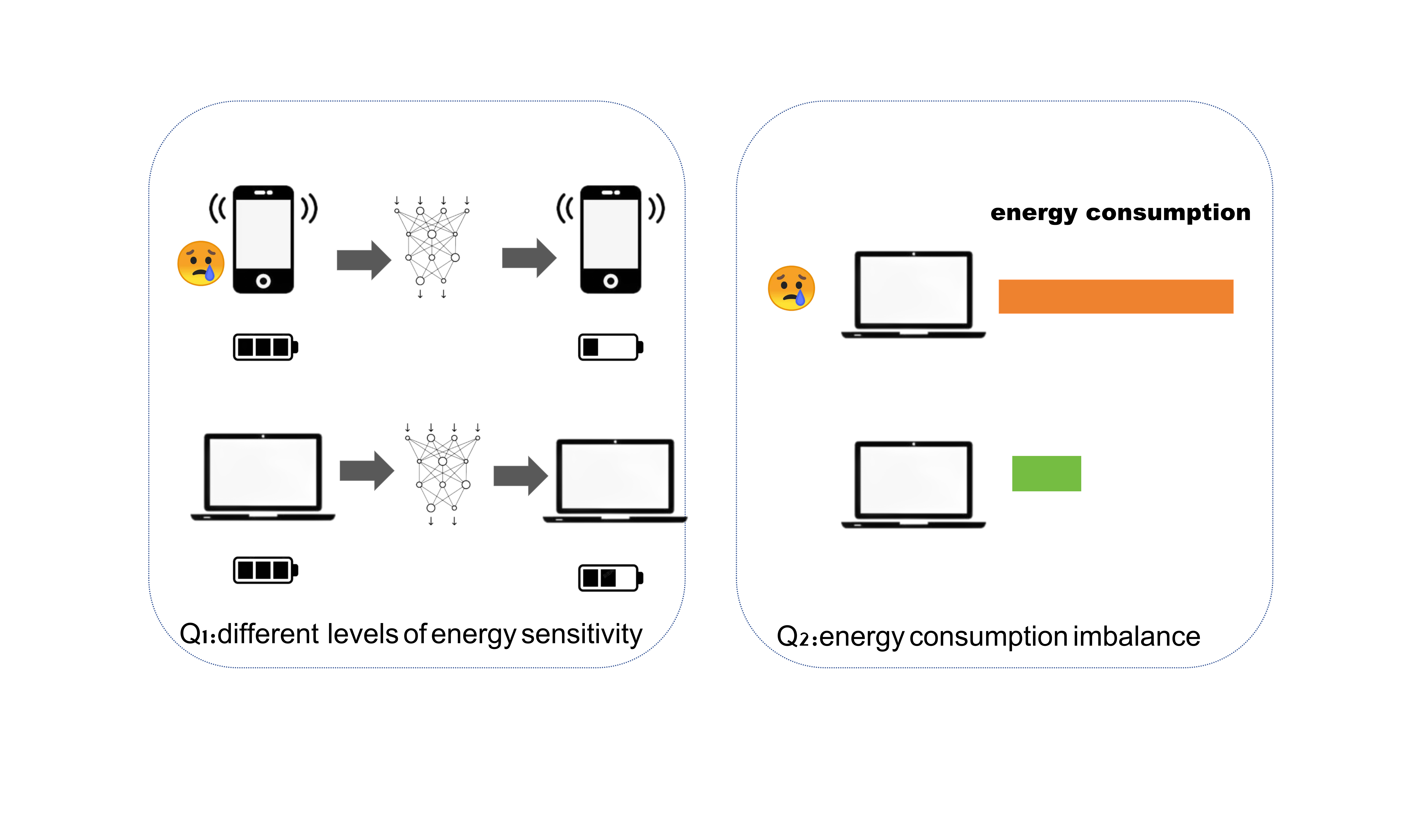}
\caption{\small The two issues in energy consumption optimization for FL: \textbf{1)} The impact of device heterogeneity on energy consumption ensitivity. \textbf{2)} The imbalance in energy consumption among devices.}
\end{figure}

In FL, leveraging heterogeneous devices is essential to improve data diversity and model accuracy. This heterogeneity leads to varying energy and time demands for tasks like model training and data communication. Ignoring these differences can result in excessive reliance on energy-hungry devices, increasing overall energy consumption and causing workload imbalances. To mitigate these issues, Cui et al. \cite{ref10} designed a scheduling strategy that uses frequency control techniques to reduce energy costs while maintaining model accuracy and performance across heterogeneous devices. Moreover, Tian et al. \cite{ref20} introduced FedRank, a novel client selection approach using a ranking-based mechanism trained with imitation learning and reinforcement learning (RL) which optimizes the selection process by considering data and system heterogeneity.

Although these works primarily address the issue of heightened overall energy consumption due to device heterogeneity, they fall short in tackling energy imbalance and the differing energy sensitivities among devices. This oversight may result in excessive energy use for certain devices or their premature withdrawal, ultimately hindering the progress of FL.

In light of this, we present the following contributions:
\begin{itemize}
\item To address client energy consumption imbalance and excessive use of energy-sensitive devices in the MEIoT environment, we recommend leveraging relative energy consumption and incorporating energy balance into the system's optimization objectives.
\item Given device heterogeneity and communication resource constraints, we apply the SLSQP algorithm for resource allocation and introduce a heuristic client selection algorithm combining clustering and utility-driven methods to enhance federated learning efficiency and reduce energy imbalance.
\item For optimal client selection during model training, we employ the BEFL framework, which uses offline heuristic learning and online RL. Experiments show that BEFL synchronizes model performance, total client energy consumption, and energy variance in dynamic training.
\end{itemize}

\section{SYSTEM MODEL AND PROBLEM FORMULATION}

\subsection{System Model}
We consider a MEIoT scenario with an edge server and multiple client devices. In each FL round, the server sends the global model to selected clients $\phi_h = \{\theta_1, \ldots, \theta_n\}$. Each client $\theta_i \in \phi_h$ trains the model on its local dataset, incurring energy $E_i^{train}$ and latency $\tau_i^{train}$. After training, the client uploads the model parameters, resulting in additional energy $E_i^{trans}$ and latency $\tau_i^{trans}$.

\subsubsection{Local training model}
The training time for device $\theta_i$ is defined as $\tau_i^{train} = \frac{I_i C_i |D_i|}{f_i} core_i$, where $C_i$, $I_i$, $|D_i|$, $f_i$, and $core_i$ represent CPU cycles per sample, local iterations, number of training samples, CPU frequency, and cores, respectively. The training energy consumption is given by $E_i^{train} = \kappa I_i C_i |D_i| f_i^2 core_i$ with $\kappa$ being the capacitance coefficient of the device.

\subsubsection{Local communication model}
The communication method using Orthogonal Frequency Division Multiple Access (OFDMA), the communication rate for device is $\theta_i$ is $r^{trans}_i = \beta_i B \log_2\left(1+\frac{g_i^2 P_i}{N_0}\right)$, where $\beta_i$ is the resource allocation ratio, $B$ is bandwidth, $g_i$ is channel gain, $P_i$ is communication power, and $N_0$ is noise power spectral density. The communication energy consumption is $E_i^{trans} = P_i \frac{G}{r^{trans}_i} = P_i \frac{G}{\beta_i B \log_2 \left(1+\frac{g_i^2 P_i}{N_0}\right)}$.

\subsubsection{Relative energy consumption model}
Total energy consumption includes both training and communication. We define relative energy consumption as $E_i^r = \frac{E_i^{train}+E_i^{trans}}{E_i^{total}\delta}$, where $E_i^{total}$ is the device's battery capacity and $\delta$ is the sensitivity coefficient.The energy consumption mentioned in this article specifically refers to the relative energy consumption under this formula.
\subsection{Problem Definition}
We address a problem within a MEIoT environment comprising an edge server and a set of client devices $\Phi = \{\theta_1, \theta_2, \ldots, \theta_n\}$. The energy consumption of each client during FL training is represented by $\Xi = \{E_1, E_2, \ldots, E_n\}$, across a total of $H$ training rounds. The set of devices selected in the $h$-th round is denoted as $\phi_h$. Our objectives are to minimize the variance in energy consumption among clients, reduce the overall energy usage, and maximize the global model's accuracy.\\
\textbf{Objective: }$\min\{ \text{Var}(\Xi), \max_{\theta_i \in \Phi} E_i, \sum_{h=1}^{H} E_{\phi_h}, L(w_g,D_{all}) \}$.
\textbf{Constraints:}
\begin{align}
E_{\phi_h} & \geq \sum_{\theta_i \in \phi_h}(E_i^{train} + E_i^{trans}), \\
T_{\phi_h} & \geq \max_{\theta_i \in \phi_h}(\tau_i^{train}+\tau_i^{trans}), \\
\sum_{\theta_i \in \phi_h}\beta_i & \leq 1, \\
\beta_i & \in [0,1], \quad \theta_i \in \phi_h, \\
\sum_{h=1}^{H} T_{\phi_h} & \leq T_{limit}.
\end{align}
These constraints ensure adequate energy allocation for training, adherence to synchronization limitations, and proper distribution of total resources within predefined limits. Due to the complexity of the issue, we will investigate heuristic scheduling methods to manage the NP-hard characteristics of the problem.

\section{Framework Design}
In this section, we present the framework's design, including the communication resource allocation strategy, the heuristic method based on offline imitation learning, and definitions related to online RL.
\begin{figure*}[hbtp]
    \centering
    \includegraphics[width=0.7\textwidth]{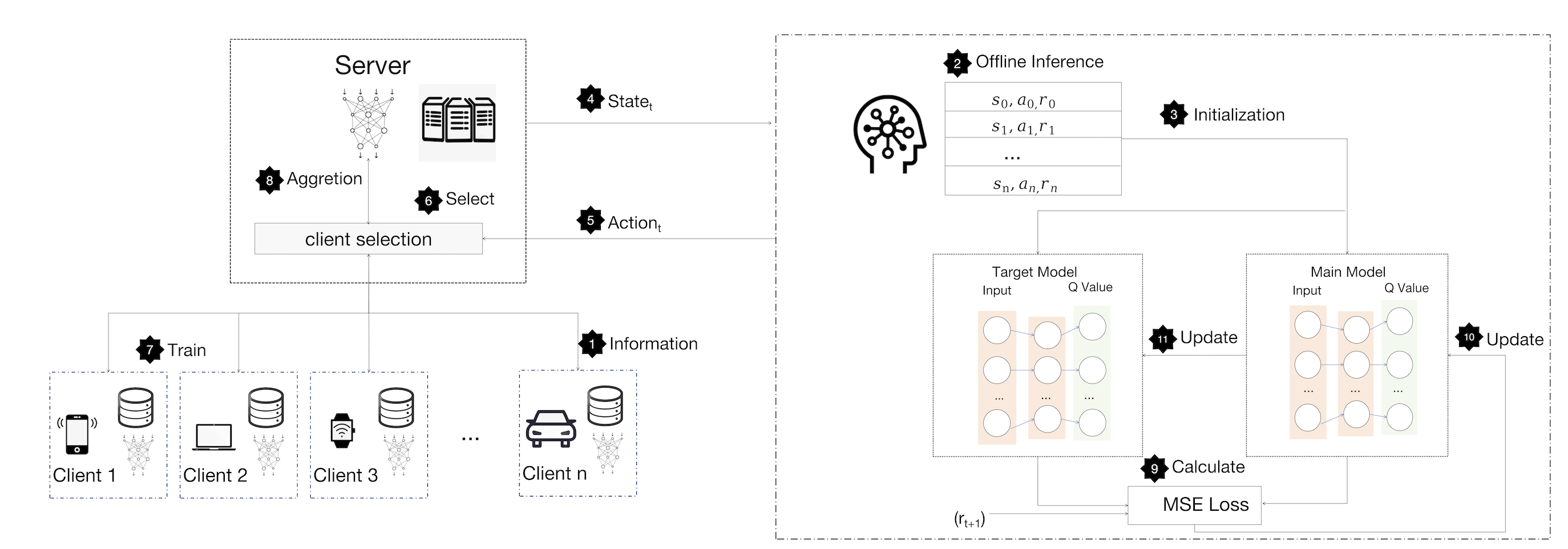} 
    \caption{\footnotesize BEFL framework overview: 1. The server retrieves hardware information from devices. 2. It simulates energy consumption and latency using energy consumption model. 3. Client selection is facilitated by a heuristic algorithm. 4. Generated state-action pairs are sent to the RL network for pre-training. 5-11. During online training, the RL agent gathers state information, selects clients, conducts local training, aggregates parameters, and updates both main and target networks. After R rounds, the target network is updated with main network parameters based on Q values and rewards.}
    \label{fig:wide_image}
\end{figure*}

\subsection{Allocate Limited Communication Resources Using the SLSQP Algorithm}
In MEIoT environments, communication resources are often limited. To optimize resource allocation, we employ the SLSQP algorithm \cite{ref16}, which minimizes communication energy consumption across selected devices during each round of federated learning.

\textbf{Objective}
$\min{\sum_{\theta_i \in \phi_h}\left(P_i\frac{G}{\beta_i B \log_2 \left(1+\frac{g_i^2 P_i}{N_0}\right)}\right)}$.

\textbf{Constraints} $\sum_{\theta_i \in \phi_h}\beta_i = 1$.
Here, $\phi_h$ denotes the set of selected clients in the $h_{th}$ round, while $\beta_i$ represents the communication resources allocated to client $\theta_i \in \phi_h$ during that round.
\subsection{Design of a Heuristic Energy-Balancing Client Selection Algorithm
}

Based on the local communication energy consumption model, devices with higher ideal energy consumption (Assuming all communication resources are obtained, $\beta_i=1)$ see a faster energy increase as allocated resource blocks decrease, compared to those with lower consumption. To mitigate this, we divide devices into two clusters: high and low ideal energy consumption. Resources are then reallocated, shifting some from low to high consumption clusters, effectively reducing total energy use and variance. To avoid over-reliance on specific devices, we introduce an efficiency function:$F(\varphi_i,E_i^{trans},E_i^{train})=\alpha^{\varphi_i}\frac{1}{E_i^{trans}+E_i^{train}}$.
Here, $\alpha < 1$ is the efficiency factor, and $\varphi_i$ denotes the selection times of device $\theta_i$ during the federated learning process. Clients will be selected based on a function. Clients selected too frequently will see a significant reduction in their objective function value, thus lowering their probability of being selected again.
\subsection{Online RL Combined with Offline Imitation Learning.}
Compared to traditional heuristic algorithms, RL algorithms are more likely to approach the optimal solution for the problem. Therefore, we adopted a framework where online RL is employed, while offline imitation learning is conducted based on the proposed heuristic algorithm.To apply RL, it must be transformed into a Markov Decision Process (MDP) by defining state space, action space, and reward function.
\textbf{State:} The state \( S^t \) at timestamp \( t \) includes:

\begin{itemize}
    \item \( s_i^T = (T_i^{t,\text{train}}, T_i^{t,\text{trans}}) \): latency of device \( \theta_i \), comprising training latency \( T_{i}^{\text{train}} \) and communication latency \( T_{i}^{\text{trans}} \).
    \item \( s_i^E = (E_i^{t,\text{train}}, E_i^{t,\text{trans}}, E_i^{t,\text{total}}) \): energy consumption, including computational \( E_i^{\text{train}} \), communication \( E_i^{\text{trans}} \), and The total consumption from FL's start to now. \( E_i^{\text{total}} \), reflecting device heterogeneity.
    \item \( s_i^D = (L_i^t, D_i^t) \): data heterogeneity, where \( L_i^t \) is training loss and \( D_i^t \) is data size.
\end{itemize}

In the offline phase or for devices that have not participated in training, we assess the status of the devices by using an energy consumption model to predict their energy consumption and latency. For active devices, we use the information from the previous round.

\textbf{Actions:} The action set \( A^t \) is an \( N \)-dimensional binary vector indicating whether device \( \theta_i \) participates in training (\( A_i^t = 1 \)) or not (\( A_i^t = 0 \)).

\textbf{Rewards:} We define three rewards—\( R_{\text{acc}} \) (test accuracy), \( R_T \) (processing latency), \( R_E \) (total energy consumption), and \( R_{\text{VarE}} \) (variance of energy consumption)—to balance multiple objectives during training rounds \( t \): \\
$[
R_t = \Delta R_t^{\text{acc}} \cdot \left( \frac{T}{T_t} \right)^{{1}(T < T_t) \alpha} \cdot \left( \frac{E}{E_t} \right)^{{1}(E < E_t) \beta} \cdot \left( \frac{V}{V_t} \right)^{{1}(V < V_t) \gamma}
]$
here, \( T, E, V \) denote penalty thresholds. If the current round's delay \( T_t \), energy \( E_t \), or variance \( V_t \) are below these thresholds, penalties are not applied. Parameters \( \alpha, \beta, \gamma \) represent the system's tolerance towards delay, energy, and balance, respectively.

To prevent large updates from deviating the model from optimal solutions, we implement two networks: a main network to predict device selections and a target network to stabilize updates:$L = \omega_{s_t,a,r,s_{t+1}} [ r_t + \gamma \sum_{i} Q^{\theta'}_i(s^{t+1}_i, a) - Q^{\theta}_i(s_i^t, a)]$.
after $( R ) rounds$, the target network updates its parameters from the main network. Once the main network infers all devices, it calculates the reward and transitions to the next state.

\section{EVALUATION}

In this section, we compare the performance differences between BEFL and the following algorithms on the CIFAR10 and MNIST datasets. 

\textbf{Random Selection:} 
(1) \textbf{FedAvg} \cite{ref3}: Basic FL framework without changes. 
(2) \textbf{FedProx} \cite{ref24}: Adjusts local iterations based on training loss for stability.
\textbf{Heuristic-based Selection:} 
(3) \textbf{AFL} \cite{ref25}: Selects devices based on model and client data. 
\textbf{Learning-based Selection:} 
(4) \textbf{Favor} \cite{ref27}: Uses accuracy to guide local weight selection. 
(5) \textbf{FlashRL} \cite{ref28}: Utilizes Double Deep Q-Learning (DDQL) to manage system and static heterogeneity in FL.

\begin{table}[ht]
    \centering
    \begin{tabular}{@{}lllcr@{}}
        \toprule
        \multirow{2}{*}{Device Type} & \multirow{2}{*}{Processor} & \multirow{2}{*}{Clock Speed (GHz)} & \multirow{2}{*}{Cores} & Price Range (USD) \\ 
                                       &                           &                               &                   &                \\ 
        \midrule
        \multirow{2}{*}{Smartphone}     & MediaTek Dimensity 9200 & 1.8-3.05          & 8                 & 3000-5000      \\ 
                                       & Qualcomm Snapdragon 8 Gen 1 & 1.8-3.0         & 8                 &                \\ 
        \midrule
        \multirow{3}{*}{Laptop}         & Intel Core i5-1240P    & 1.2-4.4           & 12                & 8000-24000     \\ 
                                       & AMD Ryzen 7 5800U      & 1.9-4.4           & 8                 &                \\ 
                                       & Intel Core i7-12700H   & 1.7-4.7           & 14                &                \\ 
        \midrule
        \multirow{2}{*}{Wearable}       & Qualcomm Snapdragon 820A & 1.6-2.15         & 6                 & 100-500        \\ 
                                       & MediaTek MT2601         & 0-1.2             & 2                 &                \\ 
        \midrule
        \multirow{2}{*}{Tablet}         & Apple A15 Bionic        & 0-3.23            & 6                 & 4000-8000      \\ 
                                       & Qualcomm Snapdragon 7c   & 0-2.5             & 8                 &                \\ 
        \bottomrule
    \end{tabular}
    \caption{Comparison of Processors in Different Devices}
    \label{tab:processor_comparison}
\end{table}

\begin{table*}[!htbp]
\centering
    \caption{\small The metrics for each client selection algorithm are based on 100 training rounds. The $round$ metric shows the rounds to reach target accuracy (IID: 99\%, Non-IID: 90\%), and $variance$ measures variation in client energy consumption. $BEFL^*$ uses a heuristic algorithm with cluster partitioning and utility-driven methods, excluding RL.}

\small 
\renewcommand{\arraystretch}{0.8} 

\begin{tabular}{c|c|cccc|cccc}
\hline
Setting & Algorithm & \multicolumn{4}{c|}{CIFAR10} & \multicolumn{4}{c|}{MNIST} \\ 
        \hline
                 &           & Acc (\%) $\uparrow$ & Energy $\downarrow$ & Variance $\downarrow$ & Latency(s) $\downarrow$ & round $\downarrow$ & Energy $\downarrow$ & Variance $\downarrow$ & Latency(s) $\downarrow$ \\ 
        \hline
IID    & Fedavg   & 51.33 & 100\% & 100\% & 2619 & 67 & 100\% & 100\% & 2771 \\ 
       & FedProx  & 51.69 & 102.4\% & 112.7\% & 2557 & 56 & 99.1\% & 75.6\% & 2751  \\ 
       & AFL      & 51.68 & 93.7\% & 75.7\% & 2358 & 76 & 81.8\% & 49.8\% & \textbf{2604} \\ 
       & FAVOR    & 51.85 & 85.5\% & 92.4\% & 2308 & 57 & 93.4\% & 64.2\% & 2814 \\ 
       & FLASHRL  & 52.05 & 95.1\% & 95.3\% & \textbf{2120} & 54  & 88.1\% & 61.3\% & 2838 \\ 
       \hline
       & $BEFL^*$ & 52.23 & 78.5\% & 31.5\% & 2393 & 54 & 82.4\% & 34.8\% & 2746 \\ 
       & BEFL     & \textbf{52.46} & \textbf{76.84\%} & \textbf{27.3\%} & 2482 & \textbf{51} & \textbf{79.8\%} & \textbf{24.6\%} & 2647 \\ 
\hline
Non-IID & Fedavg   & 41.84 & 100\% & 100\% & 3150 & 54 & 100\% & 100\% & 2830 \\ 
        & FedProx  & 42.38 & 84.0\% & 80.6\% & 3165 & 48 & 123.2\% & 108.3\% & 2935  \\ 
        & AFL      & 41.86 & 71.8\% & 69.3\% & 3360 & 63 & 89.4\% & 72.8\% & 2717 \\ 
        & FAVOR    & 41.25 & 93.3\% & 78.5\% & \textbf{2976} & 52 & 88.3\% & 64.2\% & 2932 \\ 
        & FLASHRL  & 42.54 & 81.1\% & 54.1\% & 3146 & 49 & 95.4\% & 73.4\% & \textbf{2682} \\ 
        \hline
        & $BEFL^*$ & 42.68 & 73.9\% & 31.4\% & 2393 & 51 & 82.4\% & 37.2\% & 2783 \\ 
        & BEFL     & \textbf{43.44} & \textbf{71.8\%} & \textbf{31.1\%} & 3108 & \textbf{46} & \textbf{75.2\%} & \textbf{30.4\%} & 2767 \\ 
\hline
\end{tabular}
\label{table:metrics_comparison}
\end{table*}

\subsection{Experimental Setup}
\subsubsection{Data partitioning method and client model}
For client-side training, we employ the Simple-CNN model based on the CIFAR-10 and MNIST dataset.In an Independent and Identically Distributed (\textbf{IID}) setting, we ensure an even distribution of data samples for each label across clients\cite{ref17}. Conversely, in a \textbf{Non-IID} setting, we allocate varying quantities of label data to different clients using a Dirichlet distribution with a concentration parameter of 0.5.

\subsubsection{Heterogeneous hardware settings}
We established a heterogeneous FL environment with 100 virtual clients across five mobile device types in MEIoT scenarios (Table I).The real-time CPU frequency is modeled as a normal distribution within its variation range to simulate diverse workloads.The energy sensitivity coefficient $\delta$ follows a normal distribution within the interval $[0,1]$,system's total bandwidth is 10 MHz.
\subsection{Result analysis}
\begin{figure}[H]
    \centering
    \begin{minipage}[b]{0.22\textwidth}
        \centering
        \includegraphics[width=\textwidth]{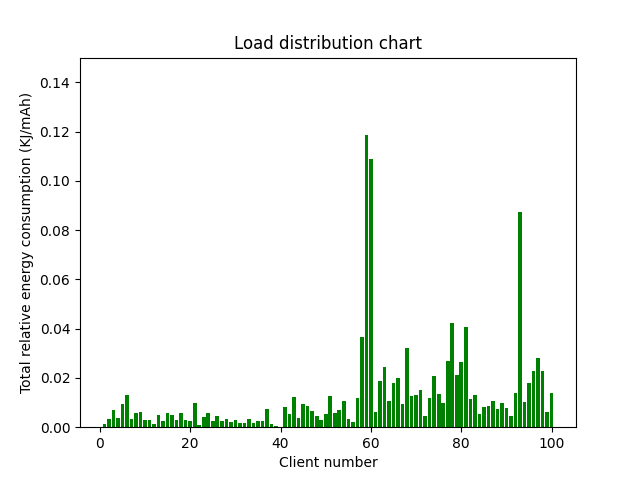}
        \subcaption{Fedavg}
    \end{minipage}
    \hfill
    \begin{minipage}[b]{0.22\textwidth}
        \centering
        \includegraphics[width=\textwidth]{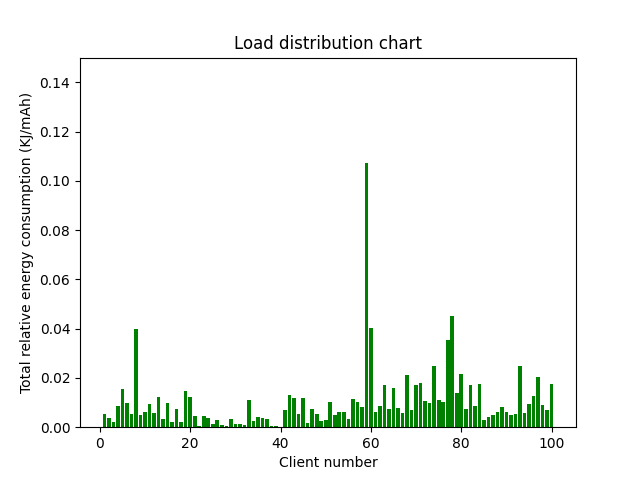}
        \subcaption{FlashRl}
    \end{minipage}
    \hfill
    \begin{minipage}[b]{0.22\textwidth}
        \centering
        \includegraphics[width=\textwidth]{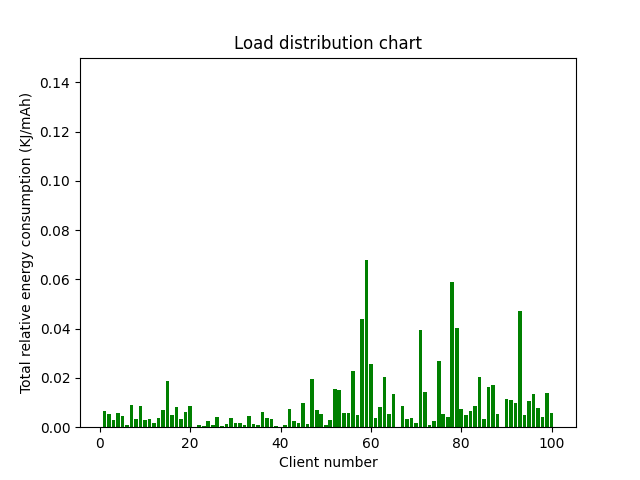}
        \subcaption{AFL}
    \end{minipage}
    \hfill
    \begin{minipage}[b]{0.22\textwidth}
        \centering
        \includegraphics[width=\textwidth]{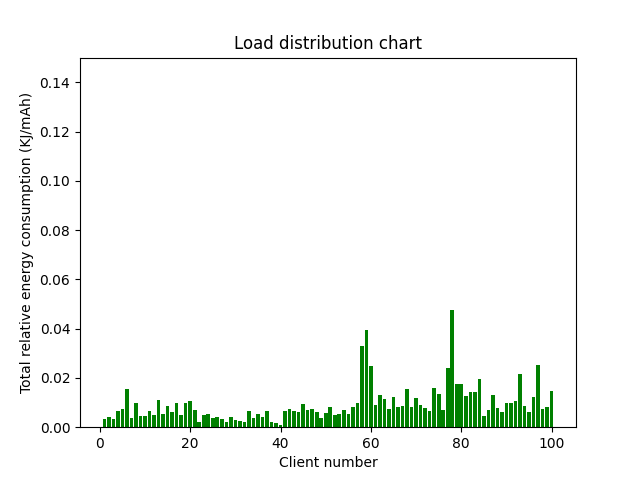}
        \subcaption{\textbf{BEFL}}
    \end{minipage}
    
    \caption{\footnotesize The relative energy consumption distribution among clients after training for 100 rounds in the Non-IID setting of the CIFAR-10 dataset.}
\end{figure}
\subsubsection{Overall result}
Tables II demonstrate that the BEFL algorithm significantly improves accuracy and balances client energy consumption without extending training time. This is due to: 1) The optimization objective of the proposed heuristic algorithm is similar to that of RL, thus serving as a warm-up effect in offline imitation learning. 2) penalties for energy consumption during online RL that maintain model accuracy while reducing energy imbalance. 
\subsubsection{Energy consumption balance performance}
Figure 3 illustrates relative energy consumption among clients after 100 rounds of FL training with various algorithms. The BEFL algorithm, leveraging a pre-trained RL model, optimizes energy use and balances consumption across clients. Compared to FedAvg, FlashRL, and AFL, BEFL reduces peak client energy consumption from 0.183 KJ/mAh, 0.112 KJ/mAh, and 0.068 KJ/mAh to 0.053 KJ/mAh, achieving reductions of 72.1\%, 54.5\%, and 22.1\%, respectively, thereby mitigating excessive load on individual clients.

\subsection{Ablation Study}
\begin{figure}[htbp]
    \centering
    \includegraphics[width=0.25\textwidth]{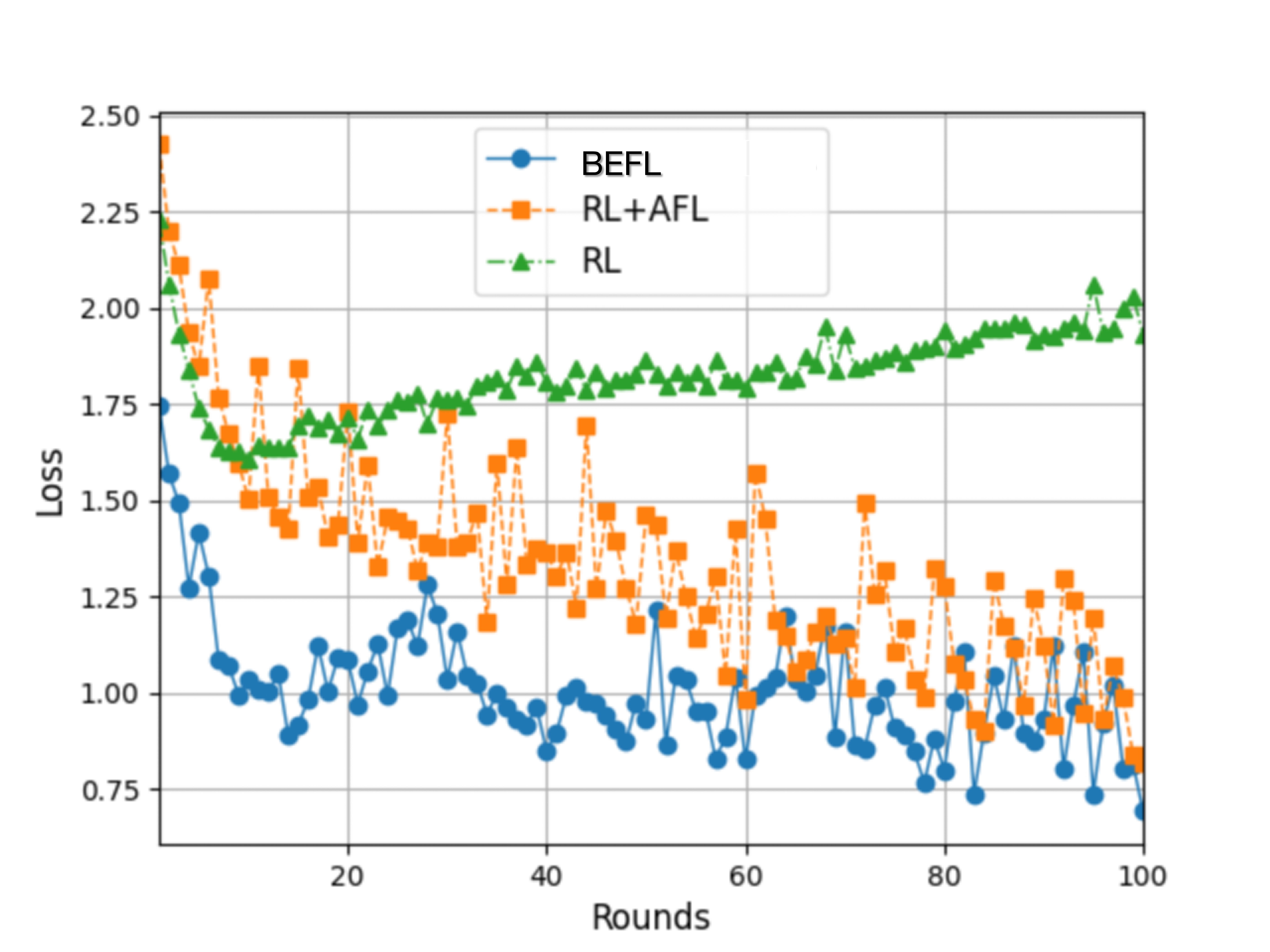}
    \caption{\footnotesize The variation of loss values for various algorithms under the Non-IID setting of CIFAR10}
    \label{fig:example}
\end{figure}

To demonstrate the role of the proposed heuristic algorithm in the warm-up phase of imitative learning, we compare three algorithms: \textbf{RL}, the RL algorithm without pre-training; \textbf{AFL+RL}, the RL algorithm pre-trained using AFL; and \textbf{BEFL}, the RL algorithm pre-trained with the proposed heuristic algorithm. As shown in Figure 4, the BEFL demonstrates superior performance and easier convergence in the early stages, likely due to the optimization objective of the proposed heuristic algorithm sharing similarities with the reward function of the RL process.
\section{CONCLUSION}
We present BEFL, a novel joint optimization framework that effectively addresses the energy consumption challenges in Federated Learning for Mobile Edge IoT. By balancing global model accuracy, total energy consumption, and energy disparities among devices, our approach enhances performance significantly—achieving a 1.6\% increase in accuracy, a 72.7\% reduction in energy disparity, and a 28.2\% decrease in overall energy consumption. These results demonstrate BEFL's potential to improve sustainability in energy-sensitive environments while maintaining model performance.

\end{document}